# Spiking based Cellular Learning Automata (SCLA) algorithm for mobile robot motion formulation

**Vahid Pashaei Rad**[1], **Vahid Azimi Rad**[2], **Saleh Valizadeh Sotubadi**[3]

[1] Master of Science, University of Tabriz, Tabriz; vahidpashaeirad@gmail.com
[2] Associate professor, University of Tabriz, Tabriz; azimirad@tabrizu.ac.ir
[3] Master of Science student, University of Tabriz; saleh1375125@gmail.com

**Abstract**
In this paper, a new method called SCLA which stands for Spiking based Cellular Learning Automata is proposed for a mobile robot to get to the target from any random initial point. The proposed method is a result of the integration of both cellular automata and spiking neural networks. The environment consists of multiple squares of the same size and the robot only observes the neighboring squares of its' current square. It should be stated that the robot only moves either up and down or right and left. The environment returns feedback to the learning automata to optimize its' decision making in the next steps resulting in cellular automata training. Simultaneously a spiking neural network is trained to implement long term improvements and reductions on the paths. The results show that the integration of both cellular automata and spiking neural network ends up in reinforcing the proper paths and training time reduction at the same time.

**Introduction**
In recent years, mobile robots have become in the center of focus due to their abilities in doing some tasks that might cause some side effects for humans. To get to such a system that is able to explore, make connections with its environment and take the best possible action, it is needed for the robot to learn. Various learning rules and algorithms are proposed for this task some of which are having desirable performances while the others might not be desired. Among these learning rules, artificial neural networks have been taken to consider more than any other rules. These networks are specifically inspired by the biological neural networks of the living species.These systems learn using several input data by extracting features of the data and learning a pattern matching the input data whether with each other or the corresponding label of any data point. In this study, a new approach for robot learning has been introduced using Spiking neural networks and cellular automata and the results are implemented on a mobile robot. Cellular Automata is a model in discrete mathematics that has been studied in various fields like mathematics, physics, theoretical biology, microstructure, and complicated adaptive systems. It is also known as homogenous structures, cellular spaces, cellular structures and repeatable arrays, [1]. It is able to simulate a number of real-world systems like biological and chemicals systems, [2]. On the other hand, Spiking neural networks are a subset of neural networks that are also known as third-generation neural networks, [3]. These networks have proven to be better than the other conventional networks in many aspects like computation ability and power consumption, [4]. In addition to neural and synaptic state, SNNs take time into consideration. In contrast to the conventional neural networks, these networks do not tend to fire in each cycle unless the membrane potential has reached a certain level known as the threshold voltage, which makes the specified neuron fire and sends signals to its subsequent neurons. This signal either strengthens or reduces the membrane potential of the subsequent neurons. The rest of the paper is as follows, first, cellular automata and spiking neural networks are defined, and then the SCLA algorithm will be described and the results will be added and finally, the results will be discussed.

**Materials and methods**
A cellular automata consists of an ordered network of cells that each one of them have one of the few possible states. There are a set of neighboring cells defined for each cell. A starting state is chosen for each cell by defining a specific state for that cell and then the new generation is generated based on a set of rules (mostly a mathematical function), [5]. The new state is chosen based on the current state of the cell and its' neighboring cells. It is assumed that the automata is in a random world choosing a specific possible action α(n) and executing it in the environment. In return, it gets β(n) as a result of the action that had previously been done by the automata that results in updating the decision making of the automata, [6]. Figure 1, illustrates the interaction between the learning automata and the surrounding environment.

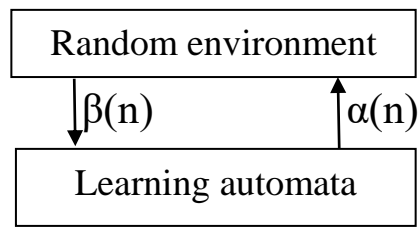

Figure1. Learning automata environment

A single learning automata is not widely used. However, with the integration of a number of single learning automata, some of the complicated problems can be solved. The integration of cellular automata with learning automata might solve the problem of having a fixed set of rules for the automata leading to learning cellular automata. A learning cellular automata of a dimension d, is shown as CLA = ($Z^d$, Q, A, N, F) the parameters are shown and defined in Table 1.

Table 1. The learning cellular automata parameters.

| Num | Learning cellular automata parameters | |
|---|---|---|
| 1 | $Z^d$ | A network in integer number space of d couples. |
| 2 | Q | A set of possible states |
| 3 | A | A set of learning automata |
| 4 | N = {$x_1, ..., x_m$} | A limited subset chosen from Z |
| 5 | F:$Q^m \rightarrow$ B | Local rule of the CLA |

The current state of the neuron which is defined by a set of differential equations is usually considered as the state of the neuron is altered by the amount of input to the neuron which leads the neuron state whether to increase and fire or to decrease and make the neuron remain at rest. The model used for defining the spiking neural network behavior is based on Izhikhevich dynamic model which is comprised of two differential equations.

$$v' = 0.04 v^2 + 5v + 140 - u - I \quad (1)$$
$$u' = a(bv - c) \quad (2)$$

In equation 1, the term v stands for the membrane potential of each neuron, u is the membrane refractory factor which leads the neuron to get to its rest state after it fires and the term I, defines the input to each neuron which either might be random or sensory input. In equation 2, the terms a, b, c, d represent the physiological parameters that have been defined in Izhikhevich model, parameter a represents the time constant of the recovery variable u which leads the membrane potential of any neuron after it has fired, parameter b shows the sensitivity of the recovery variable u to the oscillations of the membrane potential under the threshold value, the parameter c describes the after-spike reset value of the membrane potential v after the neuron has fired, the value d shows the after-spike reset of the recovery variable u [7]. Table 2 defines these parameters in a regular spiking neuron.

Table 2. Spiking neuron parameters

| Regular spiking neuron | Neuron parameters | | | |
|---|---|---|---|---|
| | a | b | c | d |
| | 0.02 | 0.2 | -65.00 | 8.00 |

The type of the learning rule used for SNNs is a reward-based learning algorithm which causes the network to learn. This reward-based system also differs from the conventional reinforcement learning methods again due to the discrete behavior of the spiking neurons. In SNN systems when a neuron fires and causes an action regardless of being right or wrong, that neuron is no longer active a moment later so tracking the neuron after a couple of moments to whether punish or reward that neuron is impossible. This issue is solved using an STDP based learning algorithm which specifies an amount of STDP parameter to the fired neurons after some times that the neurons have fired and upon that, the fired neurons get reward or punishment causing in whether strengthening or reducing the synaptic connectivity of the neurons, if the pre-synaptic neuron fires before the post-synaptic neuron then the amount of the STDP value will be increased causing in reinforcing the synaptic strength between the two neurons. Whereas, when the post-synaptic neuron fires before the pre-synaptic neuron then the value of STDP decreases causing in weakening the synaptic strength between two neurons [8].

$$c' = -c/\tau_c + STDP(\tau)\, \delta(t - t_{pre/post}) \quad (3)$$
$$\dot{s} = c \quad (4)$$

In equation 3, c is the rate of change in synaptic strength of each neuron, $t_{post}$ is the time that the post synaptic neuron fires and $t_{pre}$ is the time that the pre synaptic neuron fires. τ is the difference between $t_{post}$ and $t_{pre}$ and δ is the Dirac function. In equation 4, $\dot{s}$ is the rate of change in synaptic strength. The environment in which the simulation has been conducted is a 5*5 grid world with numbers 1 to 5 for both x and y dimensions. A fixed target is defined in a cell with coordinates of (3, 3) and a robot is defined that can move either up and down or left and right every action that it makes. Figure 2, best defines the simulation environment.

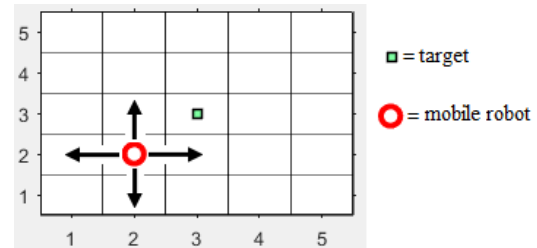

Figure 2. Defined simulation environment

The robot has four sensors and four motors that each of them are comprised of a number of neurons making an overall number of 1000 spiking neurons. These neurons are divided into two sets, the first set is called the excitatory neurons and the second set is called the inhibitory neurons in which, the sensory and motor neurons are all excitatory ones. Table 3, shows how the neurons are defined in the experiment.

Table 3. Spiking neurons definitions

| Neuron numbers | Name | Symbol |
|---|---|---|
| 1 to 100 | Upper sensory neurons | SU |
| 101 to 200 | Lower sensory neurons | SD |
| 201 to 300 | Left sensory neurons | SL |
| 301 to 400 | Right sensory neurons | SR |
| 401 to 500 | Upper motor neurons | MU |
| 501 to 600 | Lower motor neurons | MD |
| 601 to 700 | Left motor neurons | ML |
| 701 to 800 | Right motor neurons | MR |
| 801 to 850 | Upper inhibitory neurons | IU |
| 851 to 900 | Lower inhibitory neurons | ID |
| 901 to 950 | Left inhibitory neurons | IL |
| 951 to 1000 | Right inhibitory neurons | IR |

The sensory neurons are connected to the motor neurons making a 2 layer spiking neural network. There are 16 possible connections that make interactions between sensory and motor neurons. However, 8 of these connections are considered. Table 4 describes these neural connections.

Table 4. Neural connections in SNNs.

| Num | Neural connection | Symbol |
|---|---|---|
| 1 | Upper sensory neurons to upper motor neurons | SUtoMU |
| 2 | Upper sensory neurons to lower motor neurons | SUtoMD |
| 3 | Lower sensory neurons to upper motor neurons | SDtoMU |
| 4 | Lower sensory neurons to lower motor neurons | SDtoMD |
| 5 | Left sensory neurons to left motor neurons | SLtoML |
| 6 | Left sensory neurons to right motor neurons | SLtoMR |
| 7 | Right sensory neurons to right motor neurons | SRtoMR |
| 8 | Right sensory neurons to left motor neurons | SRtoML |

The distance between the robot and the target is measured. If the target is one house away from the robot's current state, then the sensory data flows through the sensory neurons by which the target has been sensed. Based on the fired neurons in a limited time step, the robot chooses one of the four possible moves and goes either up, down, left or right. If the moving direction and the sensor by which the target is sensed happen to be the same, then that connection gets strengthened and the opposite connection strength reduces, while the other two connections remain unchanged. The robot gets dopamine as a reward the moment it reaches the house in which the target has been placed. In the proposed algorithm, the probability vector does not change, unless the network sends a signal to the automata indicating that the taken action is whether good or bad.

**Results and Discussion**

The experiment was done in 30 epochs, each one being conducted through 1000 time steps. Table 5, shows the number of successful times that the robot has been able to find the target in the first 20 epoch. The experiment was also conducted using only the cellular learning automata algorithm to compare the proposed SCLA algorithm results with it. The cellular learning algorithm was also performed in 20 epochs each consisting of 1000 time steps.

Table 5. Times that the target was found using SCLA

| Number of epoch | Number of successful times that the target was found |
|---|---|
| 1 | 799 |
| 2 | 778 |
| 3 | 772 |
| 4 | 807 |
| 5 | 750 |
| 6 | 790 |
| 7 | 756 |
| 8 | 802 |
| 9 | 809 |
| 10 | 780 |
| 11 | 764 |
| 12 | 800 |
| 13 | 810 |
| 14 | 770 |
| 15 | 800 |
| 16 | 811 |
| 17 | 772 |
| 18 | 783 |
| 19 | 790 |
| 20 | 778 |

It is inferred from the experiments that, the robot gets to find the target in an average of 784 times in each 1000 time step using the proposed learning algorithm. Based on the results, it is inferred that the proposed algorithm is completely as same as expectations. The reason is that gradually the undesired connections get to lose their strength. Figure 3, shows the evolution of the synaptic connections among various sensory and motor neurons that led to a desirable result. When the desired action is done by the network, the learning automata strengthens that action and in contrast, reduces the strength of the opposite action. The same is done when a false action is done, in this case, the strength of the taken action is reduced whereas the opposite action is strengthened. In both cases, the actions on either side are unchanged. The results of the proposed algorithm help the robot take rational and optimal actions after a while and reduce any action that might cause side effects to the system.

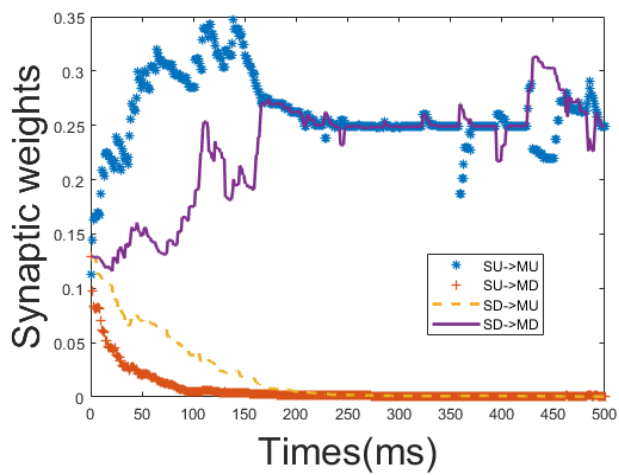

Figure 3. The evolution of different synaptic connections

## Conclusions

It is understood from the results that the integration of both learning cellular automata and the spiking neural network, end up with desirable results as the proposed learning rule optimizes the system performance and it is a good option for path planning. It was observed that in a 30 epochs simulation, the robot was able to find the target in an average of 784.2 times out of 1000 times of operation and the robot had to take at most 2 steps to reach to the target which performed better than the experiment that was done using only the cellular learning automata that resulted in an average of 115 successful time steps that the robot had found the target which meant that the robot had to make approximately 9 moves to reach to the target. It was observed that when the desired action was taken, that action would be improved whereas, the opposite action would be reduced in strength.

## Nomenclature

| | |
|---|---|
| A | a set of learning automata |
| a | time scale of the recovery variable |
| b | sensitivity of the recovery variable |
| c | after spike recovery value |
| d | after spike recovery value |
| F | local cellular automata learning rule |
| I | input of each neuron in SNNs |
| ID | lower inhibitory neurons |
| IL | left inhibitory neurons |
| IR | right inhibitory neurons |
| IU | upper inhibitory neurons |
| MD | lower motor neurons |
| ML | left motor neurons |
| MR | right motor neurons |
| MU | upper motor neurons |
| N | neighborhood vector in automata |
| Q | a set of possible states in automata |
| SD | lower sensory neurons |
| SL | left sensory neurons |
| SR | right sensory neurons |
| STDP | synaptic plasticity |
| SU | upper sensory neurons |
| SDtoMD | lower sensory to lower motor |
| SDtoMU | lower sensory to upper motor |
| SLtoML | left sensory to left motor |
| SLtoMR | left sensory to right motor |
| SRtoML | right sensory to left motor |
| SRtoMR | right sensory to right motor |
| SUtoMD | upper sensory to lower motor |
| SUtoMU | upper sensory to upper motor |
| $\dot{s}$ | rate of change in synaptic connectivity |
| t | firing time |
| $t_{post}$ | post synaptic neuron firing time |
| $t_{pre}$ | pre synaptic neuron firing time |
| u | neuron refractory parameter |
| v | neuron membrane potential |
| x | horizontal dimension |
| y | vertical dimension |
| Z | learning automata network |
| α | action of the automata in each time |
| β | environment response to automata |
| δ | Dirac delta function |
| τ | difference in firing time of 2 neurons |

## References

[1] Wolfram, S. (1983). Statistical mechanics of cellular automata. Reviews of modern physics, *55*(3), 601.

[2] Wolfram, S. (1984). Cellular automata as models of complexity.Nature, 311(5985), 419.

[3] Mass, Wolfgang (1997). "Networks of spiking neurons: The third generation of neural network models". Neural Networks. 10 (9): 1659–1671.

[4] W. Maas. Noisy spiking neurons with temporal coding have more computational power than sigmoidal neurons. In M. Mozer, M. I. Jordan, and T. Petsche, editors, advances in Neural Information Processing systems, volume 9, pages 211-217. MIT Press, Cambridge, MA, 1997.

[5] Toffoli, T., & Margolus, N. (1987). Cellular automata machines: a new environment for modeling. MIT press.

[6] Narendra, K., & Thathachar, M. A. L. (1989). Learning Automata: An Introduction. Prentice-Hall, Englewood Clis, NJ.

[7] E. M. Izhikevich "Simple model of spiking neurons," IEEE Transactions on neural networks, vol. 14, pp. 1569-1572, 2003

[8] E. M. Izhikevich, "Solving the distal reward problem through linkage of STDP and dopamine signaling" Cerebral cortex, vol. 17, pp. 2443-2452, 2007.